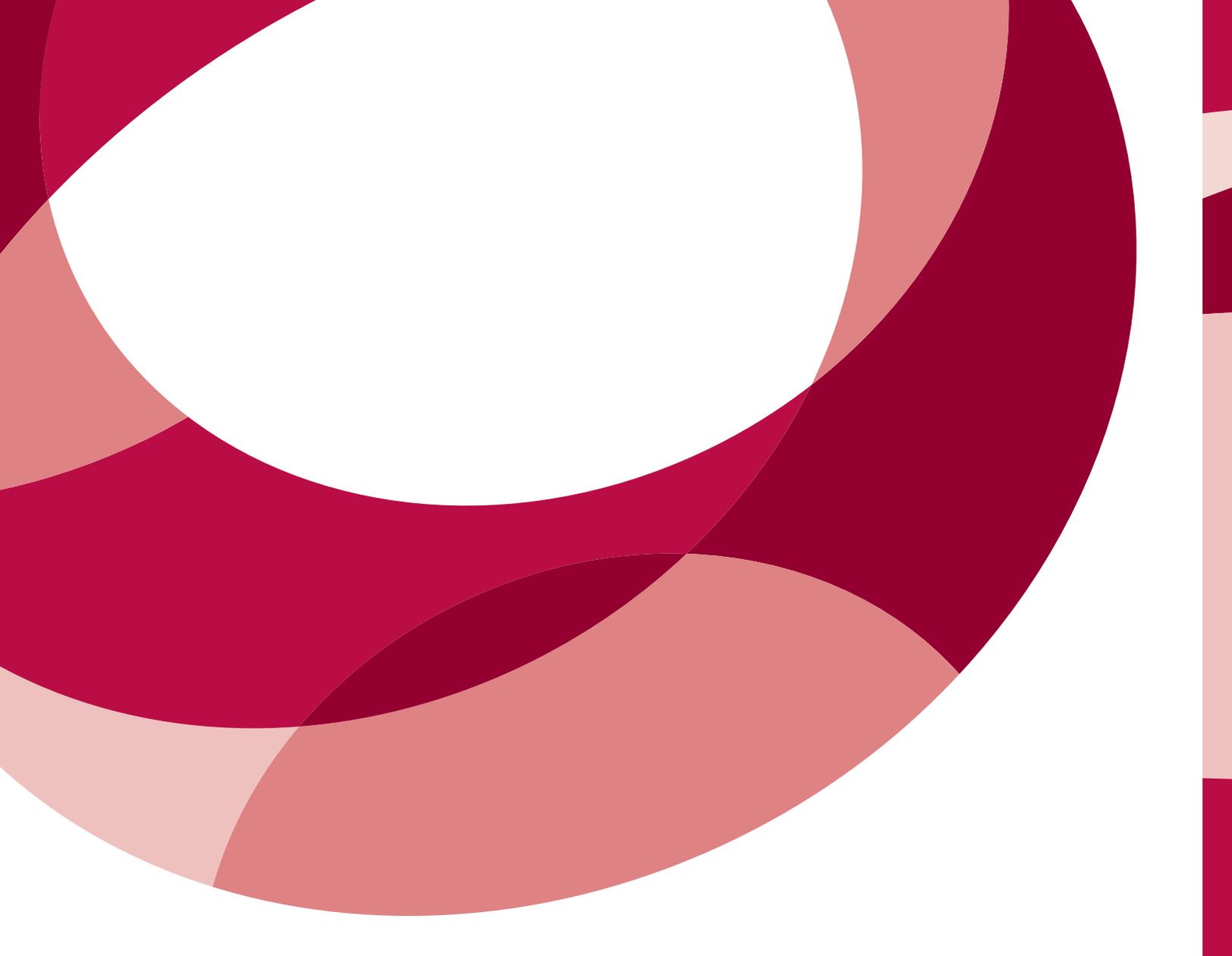

# Robotic Materials

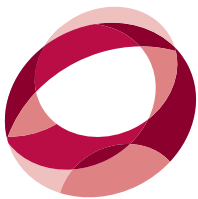

**CCC**

Computing Community Consortium

Catalyst

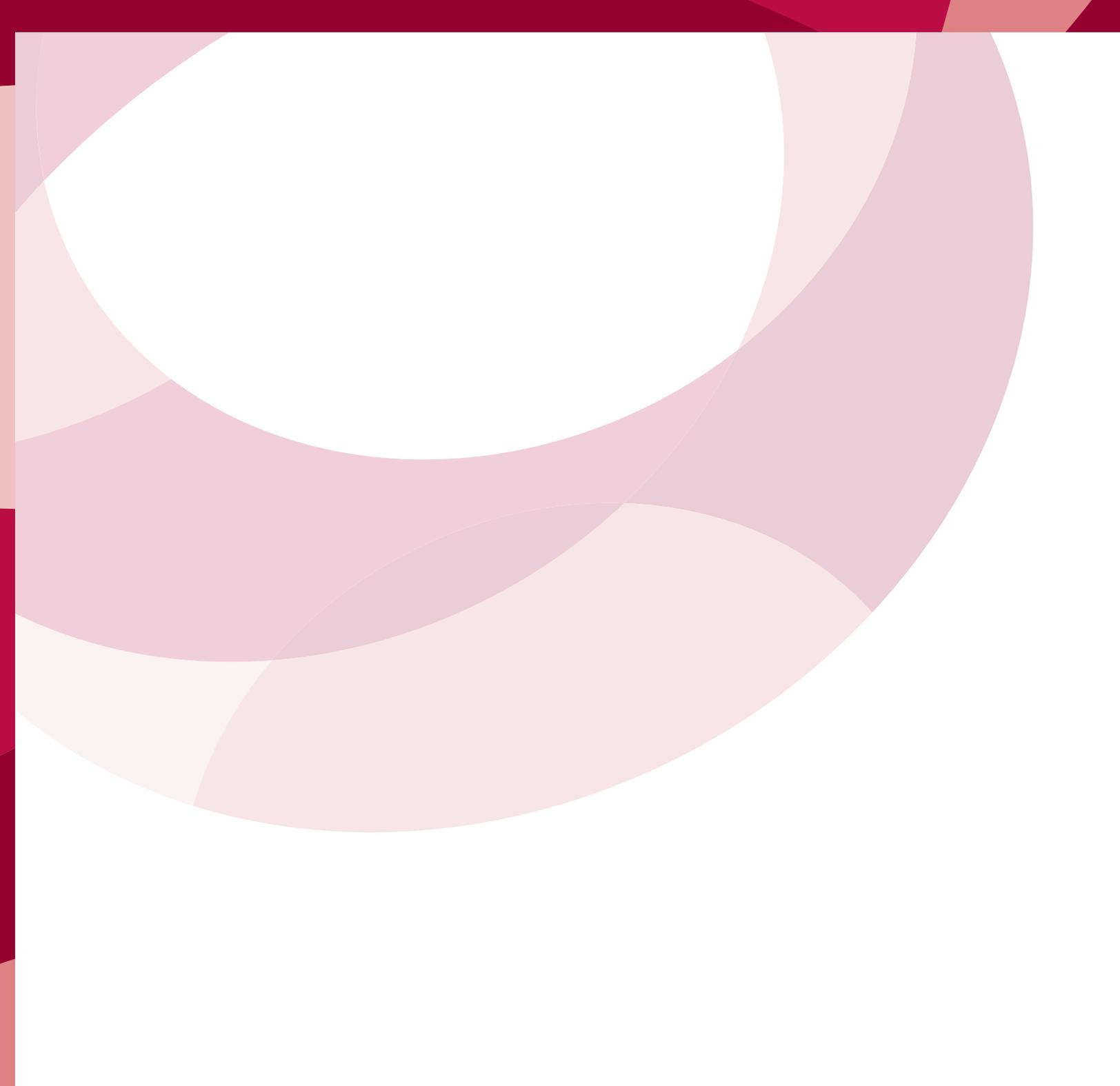

This material is based upon work supported by the National Science Foundation under Grant No.1734706. Any opinions, findings, and conclusions or recommendations expressed in this material are those of the authors and do not necessarily reflect the views of the National Science Foundation.

# Robotic Materials

Nikolaus Correll (University of Colorado Boulder), Ray Baughman (University of Texas at Dallas), Richard Voyles (Purdue University), Lining Yao (Carnegie Mellon University), and Dan Inman (University of Michigan)

November 2018

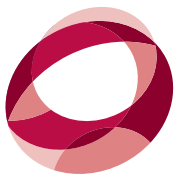

CCC
Computing Community Consortium
Catalyst

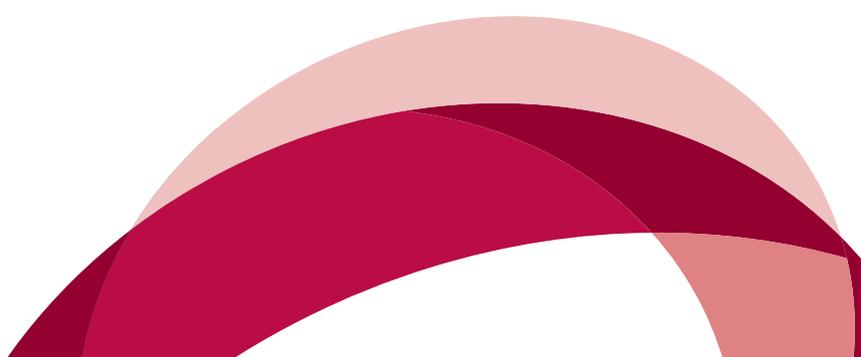

# Introduction

The Computing Community Consortium (CCC) sponsored a workshop on "Robotic Materials" in Washington, DC, that was held from April 23-24, 2018. This workshop was the second in a series of interdisciplinary workshops aimed at transforming our notion of materials to become "robotic", that is have the ability to sense and impact their environment. Results of the first workshop held from March 10-12, 2017, at the University of Colorado have been summarized in a visioning paper (Correll, 2017) and have identified the key technological challenges of "Robotic Materials", namely the ability to create smart functionality with a minimum of additional wiring by relying on wireless power and communication. The goal of this second workshop was to turn these findings into recommendations for government action. In addition to participants from the first workshop, participants for this workshop were recruited from a diverse set of stakeholders including material and computer scientists, researchers in government labs, industry, government agencies, and policymakers.

Computation will become an important part of future material systems. Computation will allow materials to analyze, change, store and communicate state in ways that are not possible using mechanical or chemical processes alone. What "computation" is and what is possibilities are, is unclear to most material scientists, while computer scientists are largely unaware of recent advances in so-called active and smart materials. For example, a nervous system is a critical part of natural material systems such as an octopus arm, a cuttlefish skin or even a bone, but it is hard to see – in the absence of an interdisciplinary definition of the following terms – what is computed and communicated. Yet, everything that can be computed can also be achieved by smart arrangement of mechanical processes, an insight rooted in the concept of Turing universality of mechanical computers. Here, a dialog and formal understanding of what is physically possible, e.g. stimulus-responsive "smart materials", and how an abstract treatment of these concepts allows for mathematical reduction might enable the creation of new materials with unprecedented functionality without requiring von Neumann architectures. This gap is currently shrinking, with computer scientists embracing neural networks and material scientists actively researching novel substrates such as memristors and other neuromorphic computing devices.

Both material and computer scientists are intrigued that biological material systems are exclusively made from cells, which is in contrast with engineered materials that are homogeneous or composites at best. Unlike conventional material systems, biological cells rely on digital information in the form of DNA that control their formation and enable their rapid evolution. It is conceivable that material engineering might reach similar sophistication, pioneered by subfields of computer science such as modular robotics, swarm robotics, social insects and amorphous computing. The latter has already begun to blend with the field of synthetic biology, creating inroads between computer scientists and chemists, and thereby material scientists.

Further pursuing these ideas will require an emphasis on interdisciplinary collaboration between chemists, engineers, and computer scientists, possibly elevating humankind to a new material age that is similarly disruptive as the leap from the stone to the plastic age. With the fast convergence of technologies that is already happening such a transition might not require an additional 300000 years, but merely a few decades.

The reminder of this report closely follows the outline of the two-day meeting, which was organized around four break-out sessions and a summary session. Participants were collecting their notes on a cloud-based "Google Slide", which serves the dual purpose of presenting to the other groups and preserving the notes online for the public. After bringing all participants onto the same page by envisioning how future material systems might look like and framing this development into a future material age that follows the bronze, iron and plastic age, the group begun to brainstorm about both civilian and military applications of next-generation materials. Serving as motivation for future scientific efforts, the groups then discussed fundamental science opportunities in both materials and their integration and how the government could support and accelerate such efforts.

# Bronze, Iron, Plastic age, how will the next material revolution look Like?

The participants generally agreed that we are observing increasing miniaturization of computation, tighter integration of functionality into volumes of decreasing size, exemplified by consumer products like the Apple watch, and a trend across all material science disciplines – from basic physics/chemistry to composite materials – to integrate computational processes inside materials. It was less clear whether this new age has already started, commonly known as the "silicon age" or whether we are on the verge of something completely novel that deserves new nomenclature such as "cooperative material systems", "emergent material systems", "dynamic materials" or "cognitive materials". Indeed, the idea is not new and has been explored in architecture ("programmable matter", Goldstein, 2005), in art ("mutant materials", Antonelli, 1995), in computer science ("amorphous computing", Abelson, 1995), in design ("radical atoms", Ishii, 2012), and robotics ("morphological computing", Pfeifer, 2006). It was duly noted that each of these words are not only individually heavily loaded, but also often difficult to define. For example, it still remains unclear what defines "robotics". We all know when we see one, but the simple requirement of something that senses, actuates, and computes is not a sufficient criterion for defining robotics. When applying these concepts to a material, it is unclear whether basic chemistry, for example when a material changes color due to temperature change, constitutes the acts of "sensing", "computing" and "actuating". Similarly, it is unclear at which point a material becomes a "system", a qualifier that was perceived to be critical by some to delineate the departure from conventional "homogeneous" materials to those with a more heterogeneous composition. Clearly, this definition is not sufficient either as heterogeneity requires a scale-based classifier to exclude alloys or even multi-cellular materials such as wood.

Informally, the idea of an "Octopus age", probably best captures the kind of material systems that might soon be enabled by the convergence of material and computer science. The Octopus' tissue tightly integrates sensors, muscles, and nervous systems to create autonomous material-like systems that can change color and shape in a non-trivial fashion. For example, an octopus arm can recognize food and aims at moving it toward the animal's mouth even after severed from the main body. A feature that is due to two thirds of the animals' neurons to be distributed in its arms. At the same time, an octopus' skin is able to generate a myriad of complex Turing-like patterns mimicking those in its environment.

Considering such examples, consensus emerged that concepts like sensing, actuation, and computation are universal in nature, a fact that is already well understood for computation where it is known as "Turing universality". In a nutshell, we can show that a Turing machine – a simple mechanism consisting of a device to read from and write binary data to an infinitely long tap – is capable of performing every possible computation. At the same, we can show that a Turing machine can be constructed from a sufficient amount of basic NAND blocks. It is therefore sufficient to show that a specific mechanism or process is functionally equivalent to a NAND block to conclude that it is capable as a building block for a general computer. This concept has begun finding its way into robotics and the material sciences via the concept of morphological computation (Hauser, 2011) and a large number of physical instantiations of logic gates using one or the other physical or chemical model. In order to reason not only about computational processes, but also about embodied behaviors, this concept needs to be extended to sensing and actuation, possibly by considering its information-theoretic tenets that might be grounded in the fact that sensing gathers and actuation alters information in one way or the other.

The image of the octopus also illustrated the properties that engineered materials with similar properties as the biological example could have. Like an animal, they could be self-healing, metabolize energy and grow by "eating", self-replicate and even perform basic computations (Reid, 2016). This functionality is, but for the most basic organisms such as slime mold that consist exclusively of identical cells, not the result of a homogenous material, but heterogeneous material systems that each consist of material systems that are of interest to this workshop. Examples here include wood, a multi-cellular "material" that is able to adapt to different loading characteristics over time, self-repairs, and stores energy, while retaining interesting structural properties even when "dead"; blood, a liquid that is able to transport energy, dissipate heat, and provide a number of sophisticated agents for self-repair; muscles, a collection of cells that is specialized to exert force in concert, and finally the brain, that performs similar function, but is focussed on information. More heterogeneous materials in which sensing, computation and actuation are tightly mixed are the tongue that combines extreme mobility with high-resolution olfactory sensors (Mengüç, 2017), the colon that is able to move solids, or the heart that is able to self-excite its actuation as a function of few external parameters.  It is likely that such basic materials could serve as the building blocks for future robots that are driven by muscles, suspended on bones and ligaments, and powered with "robot blood", motivating the original notion of "robotic materials" as describing both their internal function as well as their application.

A property that differentiates biological tissue from any engineered materials is that information on the complete system (animal, plant, fungus) is available at every cell in the form of digital information, coded by four amino acids, and serving as the mechanical blueprint for synthesizing proteins and eventually the entire organism. It really is this digital information that enables properties such as self-healing and self-replication that we find most attractive (and most difficult to understand) and it is really here that information theory and basic physics – when it comes to transcribing the sequence of amino acids into proteins – intersect. It is therefore likely that future robotic materials will have similar properties and the question remains whether we will achieve them by reverse engineering nature or simply repurposing existing biology systems such as pioneered by the field of synthetic biology, which is already emerging as one of the physical instantiations of amorphous computing (Basu, 2005), one of the inspirations behind robotic materials.

## Civilian and military applications of next-generation materials

Adding functionality by sensing, actuation, computation or a combination thereof will be beneficial for almost any material, structure or surface. Benefits might be subtle, for example by equipping surfaces with simple sensing abilities such as pressure or strain in engineered structure, or more complex, like the ability to sense bacteria to alert a consumer of the risk of food poisoning. Actuation could, for example, be used for smart packaging that can maintain a desired temperature, or change air permeability or appearance. Halloween ornaments could autonomously transform into Christmas decoration and then decompose. Furniture or pasta could be triggered to self-assemble into 3D from flat-pack pieces. The discussion also illustrated that there already exist a large number of consumer goods that make use of multi-functional materials ranging from bottles that indicate whether beer is chilled to clothing and windows that can change their thermal and ventilation properties by using a variety of physical principles. At the other end of the spectrum, the technology underlying modular robotics has already found its way into toys, and there exist prototypes of shape-changing airplane wings, and haptic surfaces that soon might become commercial products. From a military perspective, new materials could provide a strategic advantage by providing improved camouflage or protection, save energy and weight by reconfiguring, adapt to physiological conditions by adjusting the permeabilities of the fabric, and become more resilient by self-healing. Many of these properties are dual-use, that is they are equally befitting civilian applications where the same technology that enables camouflage in a military application provides opportunities for fashion design, whereas improved body armor will also help to make sporting gear and vehicles safer.

The specific applications being discussed were numerous, and will be reported not in their entirety, but to exemplify the general themes of how such products would be made and, consequently, how such products will look like. The first group are everyday-items ranging from coffee cups to clothing that take advantage of non-linear transitions in so-called "smart polymers". Here, thermochromic polymers are a simple and widely-adopted example where items change color and information appears and disappears as a function of temperature, for example to reveal information that a beer is chilled. Another example are windows that can change transparency using the electrochromic effect, clothes that can change their air permeability by opening and closing miniature latches that are activated by heat, or general shape changing composites that include liquid metal, granular jamming, or electroactive polymers that enable stiffness change by various technical means. While this first group of products derives its functionality from a single polymer, advanced manufacturing techniques will also lead to products that derive their function not only from materials, but from the specific geometry it is arranged in. For example body armor that is inspired by the Armadillo armor that consists of larger plates that can move against each other and are actuated by small muscles. Other examples in this category are shape-changing airfoils, furniture, but also camouflage skin, which will require integration of not only color changing polymer, but distributed computation and communication. Finally, there will be products that consist of "smart dust" (Warneke, 2001), smart pellets each with sensing, actuation, computation and communication abilities that can functionalize the polymer that they are integrated in. For example, particles that can sense vibrations can be integrated into a rubber tire where they collaborate to estimate the type of terrain the tire is driving on. Similarly, pellets that could induce local color change and sense their environment could self-organize to generate camouflage patterns. Like their analogue in biological cellular systems, each such pellet carries the same program and expresses different functionality as a function of its spatial-temporal location in the system it forms.

These examples also show that there exists a continuum in how specific functionality can be integrated. A bottle that indicates whether its contents are chilled can achieve this by either using a thermochromic polymer that is appropriately tuned, or as the material science community would say "programmed", or by using pellets that integrate silicon-based sensing, computation and actuation to sense the temperature, make a decision, and change the state of a few "e-Ink" pixels. While a solution using the latter approach would be much more versatile and even could display a video if so desired, there seems no application – a so-called "killer app" – that allows us to motivate the development of a general platform. Instead, for every use case of smart pellets there seem to be more direct approaches to achieve incremental functionality over the state of the art. Developing smart pellets will therefore likely result from further miniaturization of swarm and modular robotic systems as a top-down approach to realize a new class of smart materials, rather than being driven by the materials community. On the other hand, it was discussed that the two approaches, or efforts, may meet somewhere in the middle as they are advanced- when materials become machines and machines become materials.

## Fundamental science opportunities in materials and system Integration

The discussion on fundamental science opportunities was structured into "computation", "sensing and actuation" and "manufacturing". As expected, a discussion of one could not have had without the other two, highlighting the challenges of integration, which becomes a scientific challenge in its own right.

There was agreement among the participants that it is the ability to compute one way or the other that distinguishes robotic materials from conventional ones. Independently of how it is physically implemented, computation is information processing. In this sense, even a simple color-changing material computes "if temperature is larger than x, change color from a to b". A Turing machine does not only require a device to process information, but also memory. Basic material physics can be used to implement a large variety of analog computations. Here, hysteresis effects can be used to implement memory. For digital computation, the primary building block of a Turing machine, the NAND gate, doubles as memory and computational element. Either way, mechanism dynamics and chemical reactions can be reduced to mathematical equations, which allows them to be recombined to implement arbitrary computations within the limits of actual manufacturability. With different computing substrates having different advantages and drawbacks (size, speed, manufacturability, cost, flexibility), the question is therefore how to combine different means of computation (material-based, polymer electronics, silicon-based) and different architectures (digital or neuromorphic) to achieve specific computational goals.

This question is also heavily tied to the desired sensing and actuation modalities and their implementation in that these subsystems also have transfer-functions that might embody relevant signal processing. One example is the "Watts governor", a controller to regulate steam engines that takes advantage of centrifugal forces to exert forces onto a lever mechanism. The faster the engine turns, the more the lever gets activated, in turn reducing pressure to the engine, an approach that can be implemented in a large variety of basic mechanisms and complement a digital controller that might be implemented in silicon.

In addition to fundamental materials challenges on how to implement computation in materials and the design trades of the different computational substrates and combinations thereof, a different challenge is the coordination of potentially billions of computational entities in a material. At its most fundamental level, the question is "how can atoms, the basic building blocks of matter, achieve complex functionality such as a cell, an ant, or a person, out of exclusively physical interactions?" This problem can be tackled at different levels of abstraction and has received considerable interest from the computing community in the subfields of "swarm intelligence" (Bonabeau, 2001) and "swarm robotics" (Brambilla, 2013). From a materials perspective, this is highly relevant once we consider future materials to be made from smart cells that each contribute an equal amount of sensing, actuation, computation and communication ability. Which problems are most relevant is again closely tied to what kind of sensors, actuators, and finally building blocks can reasonably be manufactured now and in the future.

Here, the playing field is still wide open as we still have even not fully understood the relationship between local rules and the resulting global behavior of most social insects, despite their numbers being comparably low (when compared to the number of cells in a small mammal, for example), their communication graph being sparse (when compared to the number of synapses connecting to a single neuron in the brain) and them being observable with the eye. Making inroads toward the so-called "global-to-local problem", however, might allow us to design intelligent objects from a large number of identical building blocks, which in turn would revolutionize manufacturing. Instead of assembling systems from a large variety of processed parts, such identical building blocks could simply be molded in a desired shape. Available as a powder, they could be dissolved into liquids or rubbers, and repairing a structure would simply require to replace those parts of a structure that are actually broken.

In the meantime, we expect computation to find its way into materials by integrating conventional electronics and/or polymer electronics into composites and soft materials. Such integrations pose a myriad of design trades, for example the topology of components (amorphous or discrete), the communication channel and architecture (wireless or wired), and manufacturing challenges, for example delamination between soft and hard components, structural properties of the resulting composite, and – most importantly – how to distribute power within the material.

How to model such heterogeneous materials also poses fundamental research questions. The conventional approach is to model materials as made of finite elements whose interaction can be simulated numerically. Analytically, such systems can be represented by lumped parameter models resulting in discrete-time, discrete space equations. This is in contrast with continuous time, continuous space differential models that exist for homogenous materials. Once these systems become non-linear, for example due to integrated computers in the material that induce non-linear behavior, available analytical tools become very, very sparse.

Finally, providing energy in sufficient amounts for actuation is one of the hardest problems as batteries are infeasible in many applications, and distributing wires and creating interconnects for possibly thousands of small elements poses hard manufacturing challenges. Alternatives here are wireless power in its many variations (capacitive, inductive, solar, thermal etc.), each posing new constraints in terms of power available and what kind of geometries are feasible. There is also potential for hybrid systems consisting of completely new ways of distributing energy, for example using a liquid – akin to biological system – as a carrier and distribution system.

## Government opportunities to maintain leadership and avoid surprise in Robotic Materials

"Robotic materials" is not only a synthesis of known disciplines, but has the potential for a new field that can transform many areas of life, yet still requires distinct fundamental research before it is ready for industrial application. Advancing the field of robotic materials should therefore be in the government's interest. Examples for recent government initiatives with very similar properties are the National Robotics Initiative (NRI) and the National Network for Manufacturing Innovation (NNMI) that has lead to multiple Digital Manufacturing and Design Innovation Institutes (DMDII) across the country, among others. The NRI started with an analysis of scientific problems in which the US led and in which the US was falling behind, including identifying possible threats and opportunities from the US falling behind or the US becoming a leader. It then followed by building a community-led roadmap for research in the area for which federal investments could be leveraged for maximal benefit. The US has been leading the field since the late 90ies following a DARPA workshop (Berlin, 1997) that triggered tremendous efforts in sensor networks, amorphous computing and programmable matter, but European and Japanese researchers have developed a strong presence in swarm intelligence, swarm robotics and modular robotics.

While it might not be immediately clear what the economic opportunities of a shape-changing airplane wing and clothing and vehicles with full camouflage capabilities might be, it is pretty clear that such abilities would be a tremendous threat in the hands of adversarial powers. Also, the ability to design and manufacture systems with a tight level of integration of sensing, actuation, and computation necessitates a highly capable workforce and manufacturing environment, that – even if focussed on DoD applications – is likely to dominate also other fields of technology. One of the strongest motivators for government decision makers, in particular at state level, is job creation. Here, scientific exploration of robotic materials nurtures a workforce that is capable to innovate at the intersection of the mechanical, electrical and computer engineering, which is in increasing demand with the proliferation of embedded systems, robotics and internet-of-things devices.

A possible partner to communicate this vision to decision makers and interact with a broad set of agencies spanning from the Department of Defense (DoD) to the National Science Foundation (NSF), the Department of Agriculture and the National Aeronautics and Space Agency, is the Office of Science and Technology Policy (OSTP) in the White House and particularly the Network and Information Technology R&D (NITRD) therein. OSTP serves as a central hub for ideas and, while not equipped with any executive power by itself, it can help with introductions to the executive branch and with coordination between the various agencies.

Participants from government agencies also pointed out that unlocking new sources of money is not the primary goal. Rather, initiatives of broad nature often serve as a framework to combine existing funds into an improved, more collaborative, more transformative, narrative. For example, existing programs like NSF's Material Genome Initiative (MGI) or Designing Materials to Revolutionize and Engineer our Future (DMREF) are programs that have significant intellectual overlap with the ideas summarized in this report and are continuously evolving. Such a program might be impacted not only from the top-down, for example by lobbying directly with NSF, but also bottom-up by increasing the amount of computation in DMREF-relevant research on multi-functional materials, origami and self-assembly by the proposing scientist.

It was also noted that there are specific programs to develop common research platforms, such as was articulated after the first workshop on Robotic Materials (Correll, 2017). For example, NSF Material Centers or Emerging Frontiers in Research and Innovation (EFRI) topics might be an appropriate vehicle for developing a new innovation platform that facilitates the integration of materials and computation. Finally, Engineering Research Center (ERC) planning grants might help with structuring a larger interdisciplinary research initiative that involves not only the scientific community but also industry and workforce development.

At DoD, the Defense Advanced Research Projects Agency (DARPA) provides opportunities to develop larger programs from "seedlings", short projects of 6-9 months duration that provide evidence that a radical idea is not impossible. Within DARPA, the Microsystem Technology Office (MTO) is more systems-oriented, whereas the Defense Science Office (DSO) is more focused on fundamental research. Across DARPA, all research is centered around individual program managers, that is there exist no joint programs, making highly interdisciplinary work more difficult to pursue unless it is the specific focus of an individual program manager. This is different at other DoD agencies such as the Office of Naval Research (ONR), the Army Research Office (ARO), and the Airforce Office of Scientific Research (AFOSR), but programs in these agencies are much more driven by high-level imperatives such as an increasing focus on Artificial Intelligence in ONR that might supersede bio-inspired research. Here, robotic materials offer a pathway that connects artificial intelligence with cutting-edge material science research.

## Conclusion and recommended actions

Robotic Materials is a timely topic that provides a narrative for the convergence of disciplines that is driven by enabling technologies such as advances in smart polymers, miniaturization of computation, and new manufacturing techniques among others. Robotic materials has the potential to constitute a new material age in which man-made materials that mimic the complexity of biological tissue that includes muscles, nerves, and vascular systems become common place. The trend of materials becoming systems that integrate sensing, actuation or computation already exists, both in industry and government agencies. At the same time, robotics and artificial intelligence is becoming increasingly aware of the role of embodiment and mechanism design. Yet, the hurdles for the level of interdisciplinary collaboration that are needed are too high for truly transformative developments to happen by accident. Rather, government institutions and researchers alike should seek to push for new initiatives that require truly interdisciplinary teams with computation a required component. At the same time, researchers should seek to push the boundaries of existing programs that address the foundations of robotic materials by adding computational elements. Here, it is important to equally consider the two dominant schools of thought, those who wish to add computation by increasing material complexity, and those that wish to hope to arrive at smart materials by means of miniaturizing artificial cellular structures.

## Workshop Attendees:

| First Name | Last Name | Affiliation |
| --- | --- | --- |
| Andres | Arrieta | Purdue University |
| Christopher | Arthur | University of Virginia |
| Ray | Baughman | University of Texas at Dallas |
| Sarah | Bergbreiter | University of Maryland |
| Andy | Bernat | CRA |
| Joseph | Broz | Stanford Research Institute |
| Sandra | Chapman | Navy |
| Dick | Cheng | DARPA |
| Dan | Clingman | Boeing |
| Sandra | Corbett | CRA |
| Nikolaus | Correll | University of Colorado at Boulder |
| Dave | Cousins | Raytheon |
| Khari | Douglas | CCC |
| Ann | Drobnis | CCC |
| Richard | Han | University of Colorado Boulder |
| Peter | Harsha | CRA |
| Christoffer | Heckman | University of Colorado Boulder |
| Dan | Inman | Michigan |
| Sabrina | Jacob | CRA |
| Christoph | Keplinger | University of Colorado Boulder |
| Sang Yup | Kim | Yale University |
| Joseph | Mait | ARL |
| Tom | McKenna | ONR |
| Lynn | Millett | NAS |
| Brian | Mosley | CRA |
| Orit | Peleg | University of Colorado at Boulder |
| Ron | Pelrine | Stanford Research Institute |
| Ronald | Polcawich | DARPA |
| John | Schlueter | NSF |
| Arun | Seraphin | Senate Armed Services Committee |
| Geoffrey | Slipher | ARL |
| Henry | Sodano | Michigan |
| Cynthia | Sung | UPENN |
| Jim | Thomas | NREL |
| Richard | Vaia | AFRL |
| Richard | Voyles | Purdue |
| Ralph | Wachter | NSF |
| Shawn | Walsh | RDECOM Fellow |
| Helen | Wright | CCC |

| Shu | Yang | University of Pennsylvania |
|---|---|---|
| Lining | Yao | CMU |
| Yunseong | Nam | IonQ |
| Massoud | Pedram | University of Southern California |
| Irene | Qualters | National Science Foundation |
| Moinuddin | Qureshi | Georgia Institute of Technology |
| Robert | Rand | University of Pennsylvania |
| Martin | Roetteler | Microsoft Research |
| Neil | Ross | Dalhousie University |
| Amr | Sabry | Indiana University |
| Peter | Selinger | Dalhousie University |
| Peter | Shor | Massachusetts Institute of Technology |
| Burcin | Tamer | Computing Research Association |
| Jake | Taylor | OSTP |
| Himanshu | Thapliyal | University of Kentucky |
| Jeff | Thompson | Princeton University |
| Heather | Wright | Computing Research Association |
| Helen | Wright | Computing Community Consortium |
| Xiaodi | Wu | University of Maryland, College Park |
| Jon | Yard | University of Waterloo |
| Kathy | Yelick | Lawrence Berkeley National Laboratory |
| William | Zeng | Rigetti Computing |

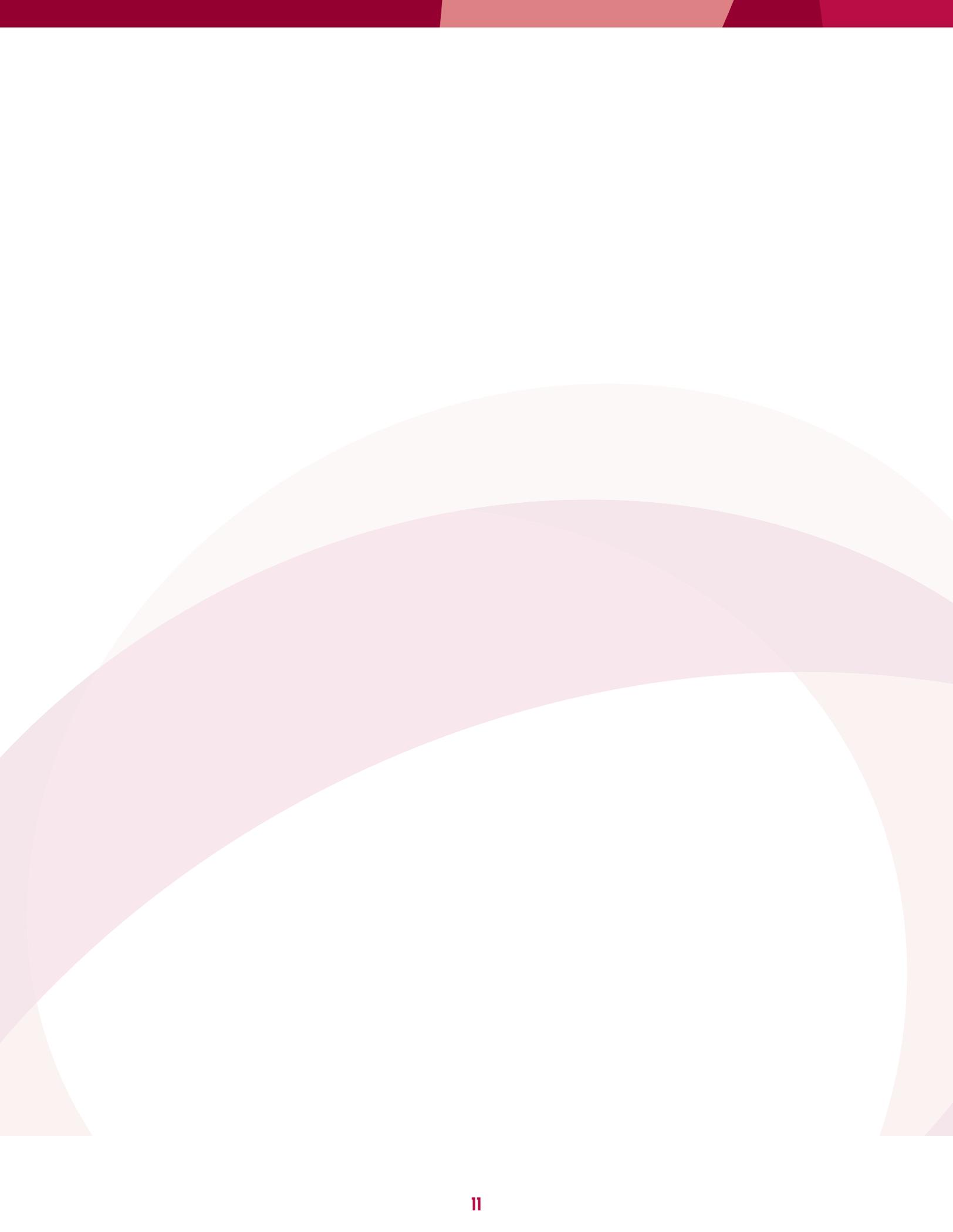

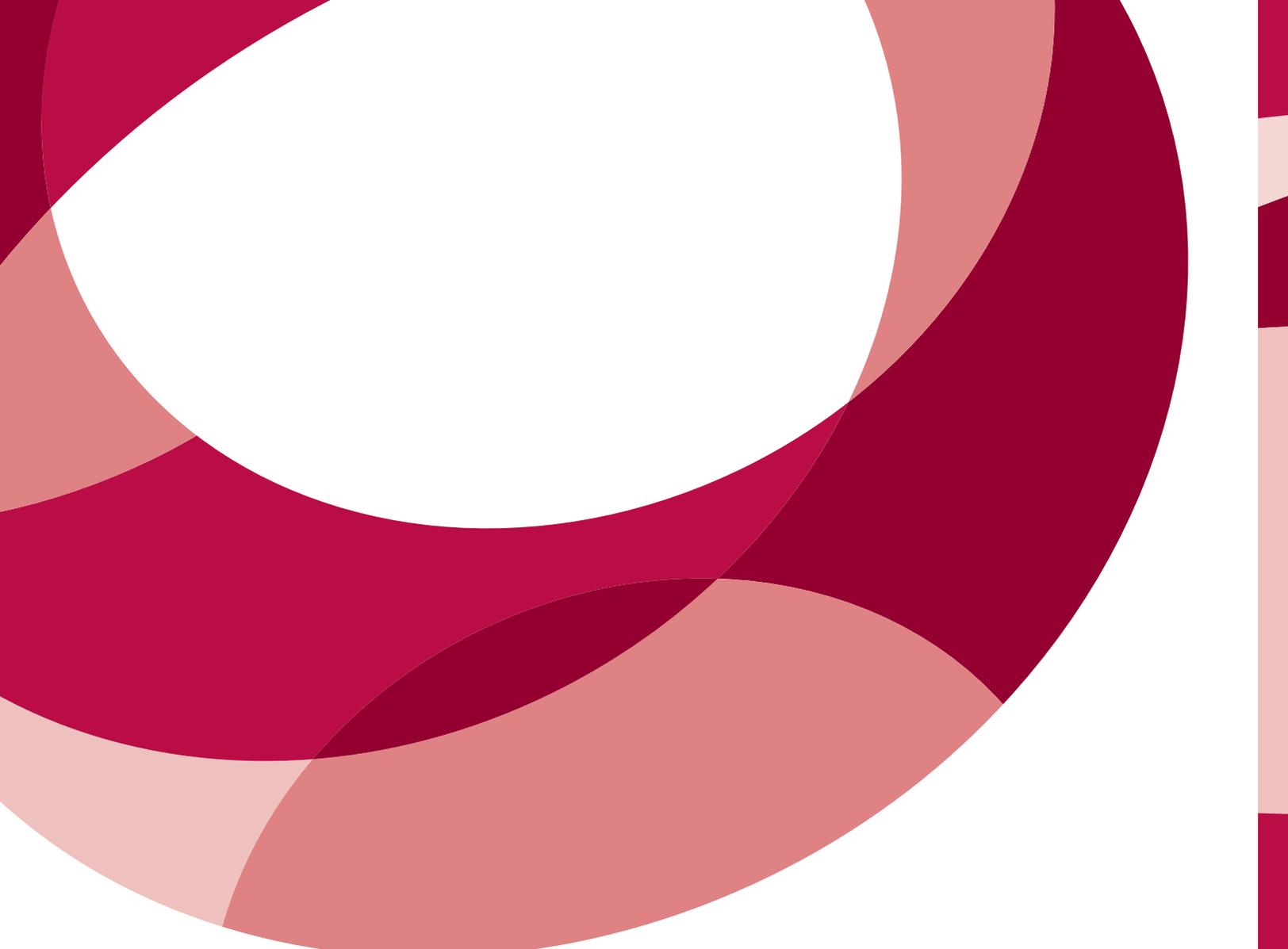

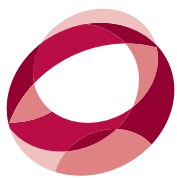

CCC
____________

**Computing Community Consortium**
Catalyst

1828 L Street, NW, Suite 800
Washington, DC 20036
P: 202 234 2111 F: 202 667 1066
www.cra.org cccinfo@cra.org